\documentclass[12pt]{article}
\usepackage{natbib}
\setcitestyle{numbers,square}
\usepackage{arxiv}
\usepackage[utf8]{inputenc} % allow utf-8 input
\usepackage[T1]{fontenc}    % use 8-bit T1 fonts
\usepackage{hyperref}       % hyperlinks
\usepackage{url} 	        % simple URL typesetting
\usepackage{booktabs} 	    % professional-quality tables
\usepackage{amsfonts}       % blackboard math symbols
\usepackage{nicefrac}       % compact symbols for 1/2, etc.
\usepackage[tracking=alltext,letterspace=0]{microtype}  % microtypography
\usepackage{lipsum}
\usepackage{graphicx}
\usepackage{epstopdf}
\usepackage{epsfig}
\usepackage{indentfirst}
\usepackage{amsmath}
\usepackage{multirow}
\usepackage{setspace}
\usepackage{amssymb}
\usepackage{xcolor}
\usepackage{threeparttable}
\usepackage{amssymb}
\usepackage{warpcol}
\usepackage{array}
\newcommand{\PreserveBackslash}[1]{\let\temp=\\#1\let\\=\temp}
\newcolumntype{C}[1]{>{\PreserveBackslash\centering}p{#1}}
\newcolumntype{R}[1]{>{\PreserveBackslash\raggedleft}p{#1}}
\newcolumntype{L}[1]{>{\PreserveBackslash\raggedright}p{#1}}

\title{WHU-NERCMS at TRECVID2021:
	Instance Search Task}

\author{Yanrui Niu\footnotemark[2]\ , Jingyao Yang\footnotemark[2]\ , Ankang Lu\footnotemark[2]\ , Baojin Huang\footnotemark[2]\ , \\
\textbf{Yue Zhang}, \textbf{Ji Huang}, \textbf{Shishi Wen}, \textbf{Dongshu Xu}, \textbf{Chao Liang\footnotemark[1]}\ , \textbf{Zhongyuan Wang\footnotemark[1]}\ ,  \textbf{Jun Chen\footnotemark[1]}\\
    \small Hubei Key Laboratory of Multimedia and Network Communication Engineering\\
	\small National Engineering Research Center for Multimedia Software, School of Computer Science, Wuhan University\\
	\small {cliang@whu.edu.cn}}

\begin{document}
    
	\maketitle
	\renewcommand{\thefootnote}{\fnsymbol{footnote}} %将脚注符号设置为fnsymbol类型，即特殊符号表示
    \footnotetext[2]{These authors contributed equally to this work.} %对应脚注[1]
    \footnotetext[1]{Corresponding author.} %对应脚注[2]

	\begin{abstract}
    	We will make a brief introduction of the experimental methods and results of the WHU-NERCMS in the TRECVID2021 in the paper. This year we participate in the automatic and interactive tasks of Instance Search (INS). For the automatic task, the retrieval target is divided into two parts, person retrieval, and action retrieval. We adopt a two-stage method including face detection and face recognition for person retrieval and two kinds of action detection methods consisting of three frame-based human-object interaction detection methods and two video-based general action detection methods for action retrieval. After that, the person retrieval results and action retrieval results are fused to initialize the result ranking lists. In addition, we make attempts to use complementary methods to further improve search performance. For interactive tasks, we test two different interaction strategies on the fusion results. We submit 4 runs for automatic and interactive tasks respectively. The introduction of each run is shown in Table \ref{Table:run}. The official evaluations show that the proposed strategies rank $1^{st}$ in both automatic and interactive tracks.
        
        \begin{table}[h!]
        \vspace{-1.0em}
        \centering
        \caption{Result of each run}
        \label{Table:run}
        \vspace{-0.5em}
        \begin{tabular}{|c|c|c|c|l|}
        \hline
        \textbf{Type}                & \textbf{Run ID} & \textbf{Relation} & \textbf{mAP}   & \multicolumn{1}{c|}{\textbf{Strategy}} \\ \hline
        \multirow{4}{*}{Automatic}   & F\_2
                & --                & \textbf{0.435} & A + F + S + R                          \\ \cline{2-5} 
                                     & F\_6            & --                & 0.418          & A + F + S                              \\ \cline{2-5} 
                                     & F\_4            & --                & 0.418          & A + F + S (w/o STE on kissing)         \\ \cline{2-5} 
                                     & F\_8            & --                & 0.395          & A + F                                  \\ \hline
        \multirow{4}{*}{Interactive} & I\_1            & F\_2 + Top-K      & \textbf{0.465} & A + F + S + R + $\text{I}_{\text{Top-K}}$                \\ \cline{2-5} 
                                     & I\_5            & F\_4 + Top-K      & 0.460          & A + F + S + $\text{I}_{\text{Top-K}}$                 \\ \cline{2-5} 
                                     & I\_3            & F\_4 + CAAF       & 0.459          & A + F + S + $\text{I}_{\text{CAAF}}$                      \\ \cline{2-5} 
                                     & I\_7            & F\_8 + CAAF       & 0.443          & A + F + $\text{I}_{\text{CAAF}}$                         \\ \hline
        \end{tabular}
        \vspace{-0.5em}
        \end{table}

        \begin{table}[h!]
        \vspace{-1.5em}
        \centering
        \caption{Description of the abbreviation in our method Introduction}
        \vspace{-0.5em}
        \begin{tabular}{|c|c|c|c|}
        \hline
        \textbf{Abbreviation} & \textbf{Description}     & \textbf{Abbreviation} & \textbf{Description} \\ \hline
        A                     & Action recognition       & R                     & Ranking aggregation  \\ \hline
        F                     & Face recognition         & $\text{I}_{\text{Top-K}}$                & Top-K feedback       \\ \hline
        S                     & Score temporal extension & $\text{I}_{\text{CAAF}}$                 & CAAF feedback        \\ \hline
        \end{tabular}
        \vspace{-1.0em}
        \end{table}
        
\end{abstract}

\section{Introduction}
\begin{figure}[t]
	\centering
	\includegraphics[width=15.5cm,height=3.5cm]{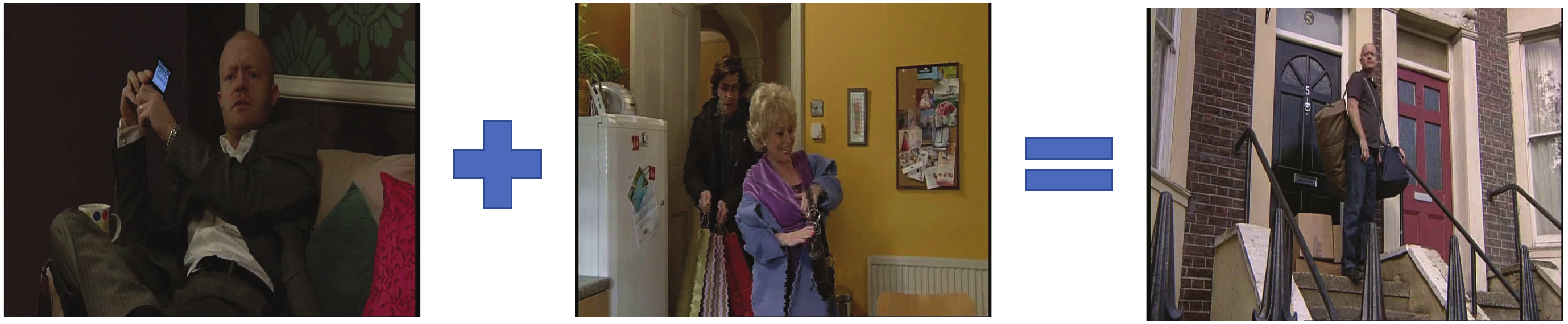}
	\caption{An example for retrieval (programme material copyrighted by BBC)}
	\label{fig:figtopic}
\end{figure}

The task of INS \cite{awad2017instance} in 2021 is retrieving specific person doing specific action as shown in Fig. \ref{fig:figtopic}, which can be represented as <person, action>. The dataset has a total of about 464 hours of videos, 471,527 shots, including 191 people, and 25 actions. The 20 topics in 2021 contain 8 actions and 6 characters. Our method is to split the person-action instance search (P-A INS) into two parts: person INS and action INS, and then fuse their separate ranking lists to generate the final results. In addition, this year we also test two interactive methods based on the fusion results.

\section{Our Method}
\begin{figure}[t]
	\centering
	\includegraphics[width=14cm,height=6.0cm]{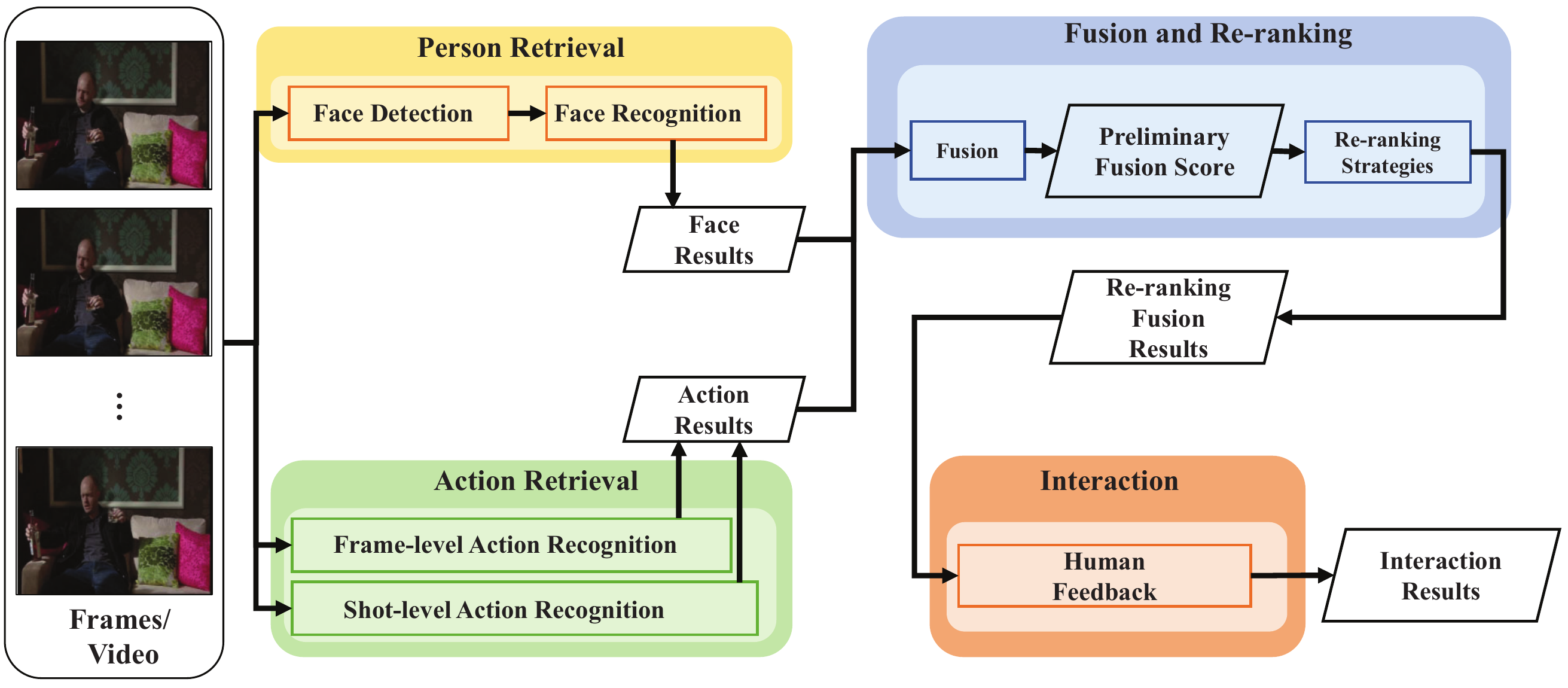}
	\caption{Our framework\quad }
	\label{fig:framework}
\end{figure}

As shown in Fig. \ref{fig:framework}, the framework we proposed for automatic task consists of four parts. The first one is the person retrieval module, which includes face detection using RetinaFace \cite{deng2019retinaface} and face recognition using ArcFace \cite{deng2019arcface}. The second part is the action retrieval module, which includes frame-level human object interaction (HOI) detection methods using PPDM \cite{liao2020ppdm}, QPIC \cite{tamura2021qpic} and ASNet \cite{chen2021reformulating}, and video-level action detection methods using TSM \cite{lin2020tsm} and ACAM \cite{ulutan2020actor}. The above two modules are used to generate the person results and the action results respectively, and the third module is the score fusion module for obtaining the ranking list by fusing the above two results. The fourth module is the re-ranking module, which adjusts the ranking result based on the scores and order of the ranking list of the fusion module.

\subsection{Person retrieval}
The person’s identity is achieved by the person retrieval module. To get the efficiency-accuracy trade-off, we resample the input videos at 5 fps. 
In the experiment, since the clarity of the official face examples is unsatisfactory, we additionally use a self-built face dataset. The official report last year showed that our face dataset performed well and improved the robustness of face recognition, so we continue to use it this year. In the person retrieval module, the first stage is the face detection, in which we choose RetinaFace. Compared with MTCNN \cite{mtcnn} used in last year, it runs faster and has higher accuracy. In particular, it improves the detection accuracy when the face rotates at a large angle, which increases the recall rate. Then in the face recognition stage, we choose ArcFace to extract features. This model proposes a new loss function to maximize the classification boundary in the annular space. Compared with the previous Center Loss \cite{Wen2016A}, it is easier to train and has a better classification effect.
After that, we compute the similarity matrix by using features from the last step and features from a self-built face dataset to get the retrieval scores.

\subsection{Action retrieval}
In this module, we use two types of methods. The first one is called frame level method, which can directly detect the interaction between people and objects in the frame, includes PPDM, QPIC and ASNet which are suitable for actions with little or slow change in time. The other type is video level method that can extract spatio-temporal information in the video, including TSM and ACAM, which deals with complex temporal actions well such as opening door and entering.

\subsubsection{Frame level}
Frame level detection methods include PPDM, QPIC and ASNet, which are used to detect HOI actions in the picture. Compared with general action detection, HOI detection focuses more on detecting the interactive actions of people and obviously objects in the picture, which can better handle some topics with HOI. Among them, PPDM is based on the CNN network structure and performs a single-stage action detection which is proved to be effective in last year. At the same time, we introduced QPIC and ASNet, which are based on Transformer and can make better use of global context information than CNN networks.

\subsubsection{Video Level}
However, there are still some actions that do not have interaction with objects or cannot be detected through a single frame. To solve these problems, we use TSM and ACAM to extract the spatiotemporal features of the videos. Among them, TSM exchanges channels between the features of neighbouring frames, which can extract video features under the cost of the 2D-CNN model. And ACAM applies I3D \cite{carreira2017quo} on the raw videos to extract features, then use an attention module to capture region of interest (RoI) for action detection, which enhances the effect of the detection with lower time cost. 

In the experiment, the score of action is generated by model results directly. For actions which are not in the pre-trained dataset, the model is used to extract the features of the action. Then, the features are compared with the features of the sample action videos, and a similarity matrix is generated to get the retrieval scores.

\subsection{Result fusion}
After obtaining the respective ranking lists of person and action, the final ranking list is obtained from the result fusion. In this part, since two methods, weight fusion and filter fusion, were verified last year and the second one shows significant advantages in the official report, all submissions this year are based on the filter fusion method. Specifically, we filter the shots containing the target person, and then finetune the action scores on these shots, which can be described as:
\begin{equation}
s_{i,j} = \boldsymbol{F}_\delta(Conf^{face}_{i}) \times Conf^{act}_{j}
\label{equ:fusion2}
\end{equation}
\begin{equation}
\boldsymbol{F}_\delta(x) = \left\{
\begin{aligned}
1, & & x \geq \delta \\
0, & & x < \delta
\end{aligned}
\right.
\label{equ:faceThreshold}
\end{equation}
where $s_{i,j}$ means the fusion result of $i$-th person and $j$-th action, $Conf^{face}_{i}$ and $Conf^{act}_{j}$ is the confidence score of $i$-th person and $j$-th action respectively. $\boldsymbol{F}_\delta({\cdot})$ is used to calculate whether the face confidence is over threshold and $\delta$ is the face threshold. For the shots with the face score less than $\delta$, the fusion score is set to 0, otherwise the action score.

By analyzing the results of each module, we find that there are still many methods to further improve the retrieval accuracy in reasonable time and computational cost. So we plus a re-ranking module additionally.

Take person INS branch as an example, it's easy to find that the face detector performs not well enough when the person has his back or side-to-side towards the camera. Besides, during one shot, the face may be temporarily obscured due to the movement of people or other objects. Similarly, the action also suffers from failed detection when objects which are part of action are temporarily invisible due to person movements or camera scope. 
So we propose an inter-frame detection extension (IDE) \cite{yang2021spatio}, which can fill in the gaps among shots due to detection failures. Firstly it will scan the results of a shot and find the failure slices between two effective detections that has the same person id or action id. Then the confidence score for the face detection box can be complemented as: 
\begin{equation}
Conf^{k}_{i} = \frac{n}{m + n} \times Conf^{k - m}_{i} + \frac{m}{m + n} \times Conf^{k + n}_{i}
\label{equ:IDE}
\end{equation}
where $Conf^{k}_{i}$ is the confidence score of $i$-th person in the $k$-th keyframe which failed to detect. $Conf^{k - m}_{i}$ and $Conf^{k + n}_{i}$ are the scores of its neighbours, the $(k - m)$-th and $(k + n)$-th keyframes. We use linear interpolation to calculate the confidence scores and the positions of face detection boxes.

However, directly fusing the results of person and action module like Eq. (\ref{equ:fusion2}) is suboptimal, because of 
$identity \ inconsistency \ problem$ (IIP), which means that there may be more than one person in one shot, and the person identity of detected face and action from two branches may not accordant.
Hence, we propose identity consistency verification (ICV) \cite{yang2021spatio} to solve the problem. Since both action and person module can provide the box of person, we can calculate the intersection over union (IoU) of the action box and face box. The assumption is that the action box and face box will have a higher possibility of belonging to the same owner if they have higher IoU, and Eq. (\ref{equ:fusion2}) can be rewritten as:
\begin{equation}
s_{i,j} = c_{i,j} \times \boldsymbol{F}_\delta(Conf^{face}_{i}) \times Conf^{act}_{j}
\label{equ:fusion_ICV}
\end{equation}
\begin{equation}
c_{i,j} = \frac{\boldsymbol{Area}(Box^{face}_{i} \bigcap Box^{act}_{j})}{\boldsymbol{Area}(Box^{face}_{i})}
\label{equ:ICV}
\end{equation}
where {$s_{i,j}$ is the score of $i$-th person and $j$-th action in the keyframe. $c_{i,j}$ is the score of ICV result which is added as weight of origin result. $Box^{face}_{i}$ and $Box^{act}_{j}$ is the box of $i$-th person and $j$-th action respectively. We use the percentage of the overlapping area in the face area as $c_{i,j}$.}

\subsection{Re-ranking}
Additionally, we found that some actions have temporal continuity and can last more than one shot, so we propose score temporal expansion (STE) \cite{yang2021spatio}, which adjusts the fusion score of the special shots by fusing the scores of neighbour shots.
In the experiment, we set the diffusion direction from higher confidence shots to lower ones, and calculate the score according to the difference between the score of two shots and the distance between two shots:
\begin{equation}
s^{k\_ste}_{i,j} = s^{k\_ori}_{i,j} + \theta\sum_{-p<m<p} \boldsymbol{F}_{dis}(m) \times \boldsymbol \max(s^{(k+m)\_ori}_{i,j} - s^{k\_ori}_{i,j},  0)
\label{equ:STE2}
\end{equation}
\begin{equation}
\boldsymbol{F}_{dis}(m) =  e^{-\frac{m^{2} }{\sigma} } 
\label{equ:dis_weight}
\end{equation}
where $s^{k\_ste}_{i,j}$ is the revised score of $i$-th person and $j$-th action  in the $k$-th shot after STE stage, $s^{k\_ori}_{i,j}$ is original score, and ${F}_{dis}(\cdot)$ is the distance weight decaying with the length between two shots. Using maximum function to limit direction and hyperparameter $\theta$ and $\sigma$ to change size.

The Ranking Aggregation (RA) strategy can obtain better results than any of the original ranking lists by fusing the results. We apply the method in \cite{mohammadi2020ensemble}, which is based on the Half-Quadratic (HQ) theory and optimizes the results by minimizing the distance between the result sorting and all input sorting. Different from the traditional sorting fusion algorithm, this algorithm uses the HQ function instead of Euclidean distance to measure the distance, so that the final result is less affected by abnormal sorting. At the same time, based on the HQ theory, the algorithm transforms the distance minimization problem into an iterative problem. Through continuous iteration until the weight vector converges, the weight of each order is calculated and merged to obtain the final order. The objective function of the algorithm is as follows: 
\begin{equation}
\min_{R^{*} ,\alpha} J(R^{*},\alpha) =\sum_{m=1}^{M} \alpha _{m}\left \| R^{m}-R^{*} \right \|_{2}^{2} +\psi(\alpha _{m})       \label{ensemble}
\end{equation}
Among them, $M$ is the number of sorted lists, $R^m$ is the sort of the $m$th sorted list, $R^*$ is the total sort of the aggregated $M$ sorted lists, $\alpha \in R^{M} $ is the HQ auxiliary variable, and $\psi(\cdot)$ is the conjugate convex function of the HQ function $g(\cdot)$.

\subsection{Interaction}
In the interaction module, we used two different interaction methods called Top-K Feedback and confidence-aware activate feedback (CAAF) \cite{zhang2021confidence}, both of which are proved to be effective.

\subsubsection{Top-K Feedback}
For the first method, we used a simple rearrangement method that is directly labeling the top-k shots in the ranking list firstly, then promoting the marked positive samples to the top of the list and putting the negative examples to the end. Since this method does not introduce the wrong samples to the top of the list, the experimental results show that the method performs well. Besides, we have many labeling strategy for different conditions. 

\begin{itemize}
	\item Only label positive examples: Successively check top-K examples and only label positive ones, then put them to the top of ranking list.
	\item Only label negative examples: Successively check top-K examples and only label negative ones, then put them to the bottom of ranking list.
	\item Label positive and negative examples: Successively check top-K examples and label both positive and negative ones, the positive examples of that are pull up to the top while the negative examples of that are push down to the bottom.
\end{itemize}
we can only label positive or negative examples for speeding up or label both of them for fine-grained labels.

\subsubsection{CAAF}
Considering that the features of similar actions are closer in feature space, while different actions are farther away, we apply CAAF. This method can dynamically recommend elements according to the current ranking list and the interaction condition of the elements. CAAF chooses the average of the top $A (A>10)$ results as probe. Each elements of probe and gallery has a confidence score represented by ${v}$ and ranking score represented by ${f}$. CAAF chooses high quality samples according to ${v}$, and changes the ranking list according to ${f}$. Human feedback can change ${v}$, and CAAF can correspondingly refresh ${f}$ for all elements. The optimization goal is shown below:
\begin{equation}
\mathop{\min}_{ \boldsymbol{f}, \boldsymbol{v}} ~\mathcal{E}( \boldsymbol{f}, \boldsymbol{v}) = \mathcal{L}( \boldsymbol{f}, \boldsymbol{v}) + \mathcal{R}(\boldsymbol{v}) 
\label{equ:CAAF}
\end{equation}
Where $i$ and $j$ means the $i$-th and $j$-th element in the set $X$ which includes both probe and gallery elements. $\mathcal{L}( \boldsymbol{f}, \boldsymbol{v})$ is calculate by $\mathcal{L}( \boldsymbol{f}, \boldsymbol{v}) = \frac{1}{m^2} \sum_{i,j}(v_i+v_j)(l_{ij}-\beta)$, and $\mathcal{R}(\boldsymbol{v})$ is a penalty for limit ${v}$. $l_{ij}$ and $\beta)$ are pairwise loss and loss threshold.

\section{Analysis}
\footnotetext[3]{F\_8 is the abbreviation of the submission ID F\_M\_A\_B\_WHU\_NERCMS.21\_8, for the interactive submissions, I\_7 is the abbreviation of I\_M\_A\_B\_WHU\_NERCMS.21\_7, we will use the similar form below.}
Table. \ref{Table:run} shows the mAP and method used in the all of our submission, so we analyze the effectiveness of each technique by comparing the difference between them. By only fusing the score of action and person module and using PPDM, QPIC, TSM and ACAM for action module, F\_8$\footnotemark[3]$ is our automatic baseline that achieves $0.395$ mAP. F\_6 and F\_4 are equipped with Score Extension strategy, both of which get the mAP boost of $0.023$ mAP . The difference of them is the choice of topics, for we only use Score Extension on the action which may last for more than one shot. However, the constancy of some action are alterable according to the context. And the result reveals that Score Extension performs not well enough on these actions. F\_2 is added with Ranking Aggregation strategy, which contributes $0.017$ mAP. We use PPDM, QPIC and ASNet to generate result respectively, and after that, use the ensemble method in \cite{mohammadi2020ensemble} to fuse the above three score to generate a better ranking list. So it is the best automatic run among our submissions. 

For interactive task, I\_7, the baseline of our interactive submission, uses CAAF strategy on I\_8. With Query Expansion strategy, CAAF can find the valuable example for human interaction so that it can get $0.048$ mAP increase. I\_5 and I\_3 apply Top-K Feedback and CAAF on I\_4 and got $0.042$ and $0.041$ mAP increase respectively. Since Top-K Feedback can provide more interactive examples and CAAF can provide more valuable examples, these submissions achieve the comparable result. I\_1 is generated by Top-K Feedback strategy from I\_2, due to higher automatic result, it is the best interactive run among our submissions and up $0.030$ mAP.
\begin{figure}[t]
	\centering
	\includegraphics[width=16cm,height=6.5cm]{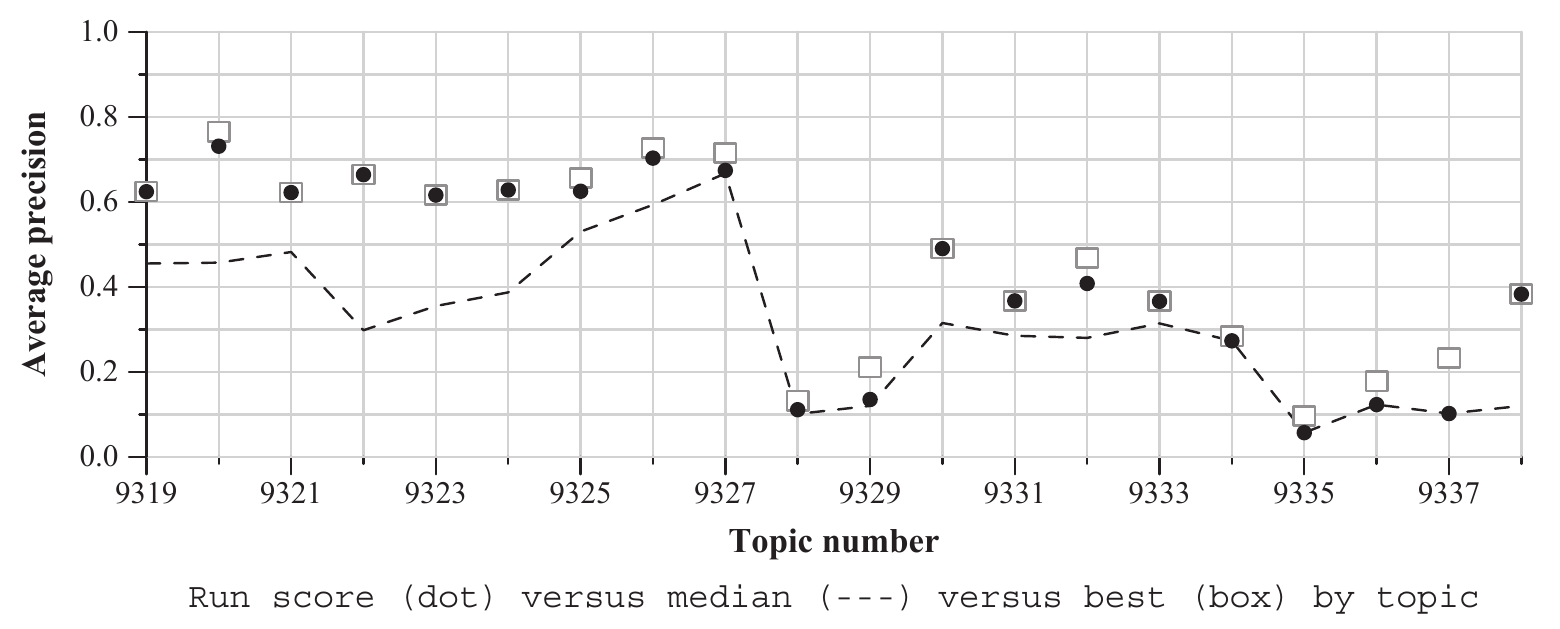}
	\caption{Comparison of automatic task submission}
	\label{fig:comp_auto}
\end{figure}
\begin{figure}[t]
	\centering
	\includegraphics[width=16cm,height=6.5cm]{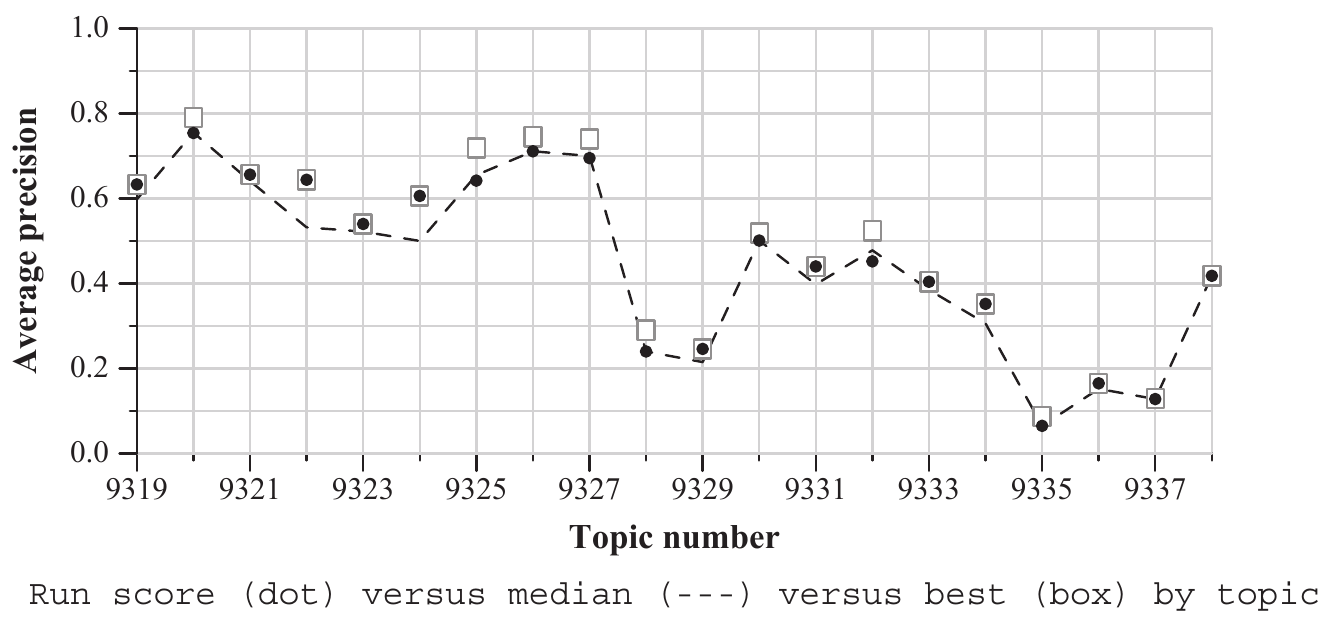}
	\caption{Comparison of interactive task submission}
	\label{fig:comp_inter}
\end{figure}

Fig. \ref{fig:comp_auto} and Fig. \ref{fig:comp_inter} shows the automatic and interactive result of I\_2 and I\_1 in all topic this year. The 'dot', 'box' and '-' are our score, the best score and the median score of each submissions from all entrants. It is obvious that our best submission achieves top-1 at quite a lot of topics (9 for automatic task and 12 for interactive task). For many topics with obvious HOI, our method exhibits superior accuracy, due to the advanced frame level detector. However, in some topics, such as 9328 (<'Max', 'carrying bag'>) and 9329 (<'Peggy', 'carrying bag'>), our method performs not well on them. By reviewing the ranking lists, we find that our model prefer to mistake 'bag' for 'belt', which is similar with necklace and other jewelry and hence generates wrong results. For 9335 (<'Bradley', 'open door enter'>) and 9336 (<'Pat', 'open door enter'>), the action changes a lot over time and is confusing compared with 'open door leave', 'open door stand' and so on. So our method does not perform well on these actions.

\section{Conclusion}
Through the INS task in TRECVID 2021, we conduct extensive experiments for our framework. By using advanced models and re-ranking strategies, our submission achieves the $1^{st}$ place for automatic and interactive searches.

\section*{Acknowledgement}
This work is supported by National Nature Science Foundation of China (No. U1903214, 61876135, 62071338, 61862015), National Nature Science Foundation of Hubei Province (2019CFB472) and Hubei Province Technological Innovation Major Project (2018AAA062). The numerical calculations in this paper have been done on the supercomputing system in the Supercomputing Center of Wuhan University.

\bibliographystyle{unsrt}
%\bibliography{trecvid2021}

\end{document}